%% file: ms.tex
\documentclass[runningheads]{llncs}

\usepackage{graphicx}
\usepackage{comment}
\usepackage{amsmath,amssymb}
\usepackage{color}
\usepackage{url}
\usepackage{hyperref}

\newif\ifreview
\reviewfalse

\ifreview
	\usepackage{lineno}

	\linenumbers
\fi

\begin{document}

%%%%%%%%%%%%%%%%%%%%% Add submission id, track, and title. %%%%%%%%%%%%%%%%%%%%%

% Insert your submission number here
\def\SubNumber{029}

% Choose one track by uncommenting one of the following lines  
% \def\GCPRTrack{Regular Track}
% \def\GCPRTrack{Track: Computer vision systems and applications}
% \def\GCPRTrack{Track: Pattern recognition in the life and natural sciences}
% \def\GCPRTrack{Track: Photogrammetry and remote sensing}
% \def\GCPRTrack{Track: Robot vision}
% \def\GCPRTrack{Track: DAGM Young Researcher Forum}

% Replace with your title
\title{TetraPackNet: Four-Corner-Based Object Detection in Logistics Use-Cases}
% You can use \thanks for acknowledgment. Do not add any acknowledgment to the draft 
% version that is used for the review process.  
%\title{Title\thanks{XXX}}

\ifreview
	% ANONYMOUS SUBMISSION FOR REVIEW
	% DO NOT MODIFY these for the draft version that is used for the review process.
	\titlerunning{DAGM GCPR 2021 Submission \SubNumber{}. CONFIDENTIAL REVIEW COPY.}
	\authorrunning{DAGM GCPR 2021 Submission \SubNumber{}. CONFIDENTIAL REVIEW COPY.}
	\author{DAGM GCPR 2021 - \GCPRTrack{}}
	\institute{Paper ID \SubNumber}
\else

	\author{Laura D\"orr\inst{1} \and
	Felix Brandt\inst{1} \and
	Alexander Naumann\inst{1} \and
	Martin Pouls\inst{1}}
	
	\authorrunning{L. D\"orr et al.}
	
	\institute{FZI Forschungszentrum Informatik\\Haid-und-Neu Straße 10-14, 76131 Karlsruhe, Germany \\
	\email{doerr@fzi.de}}
\fi

\maketitle              % typeset the header of the contribution

\input{abstract.tex}

\input{1-introduction.tex}
\input{2-relatedwork.tex}
\input{3-method.tex}
\input{4-use-case-data.tex}

\input{5-experiments.tex}
\input{6-conclusion.tex}

\bibliographystyle{splncs04}
\bibliography{content}

\end{document}

%% file: abstract.tex
\begin{abstract}
	While common image object detection tasks focus on bounding boxes or segmentation masks as object representations,
	we consider the problem of finding objects based on four arbitrary vertices.
	We propose a novel model, named TetraPackNet, to tackle this problem.
	TetraPackNet is based on CornerNet and uses similar algorithms and ideas.
	It is designated for applications requiring high-accuracy detection of regularly shaped objects, which is the case in the logistics use-case of packaging structure recognition.
	We evaluate our model on our specific real-world dataset for this use-case.
	Baselined against a previous solution, consisting of an instance segmentation model and adequate post-processing, TetraPackNet achieves superior results (9\% higher in accuracy) in the sub-task of four-corner based transport unit side detection.
\end{abstract}

%% file: 1-introduction.tex
\section{Introduction}
\label{sec:introduction}

While common image recognition tasks like object detection, semantic segmentation or instance segmentation are frequently investigated in literature, some applications could greatly benefit from more specialized approaches.
In this work, we investigate how such specialized algorithms and neural network designs can improve the performance of visual recognition systems.
For this purpose, we consider the use-case of logistics packaging structure recognition.

%Standardizing image recognition tasks and working on benchmark datasets to evaluate emerging methods and approaches is obviously a scientific necessity, allowing for comparability and meaningful evaluations or rankings of research contents.
%Further, in many, if not most, image recognition scenarios and applications, standardized approaches can cover specific requirements and specifications adequately. 
%In some other cases, however, it might be beneficial, or even inevitable, to incorporate object or segment representations differing from bounding boxes or pixel-based masks.

\begin{figure}[htb]
	\centering
	\includegraphics[trim=0 20 0 150,clip,width=0.36\textwidth]{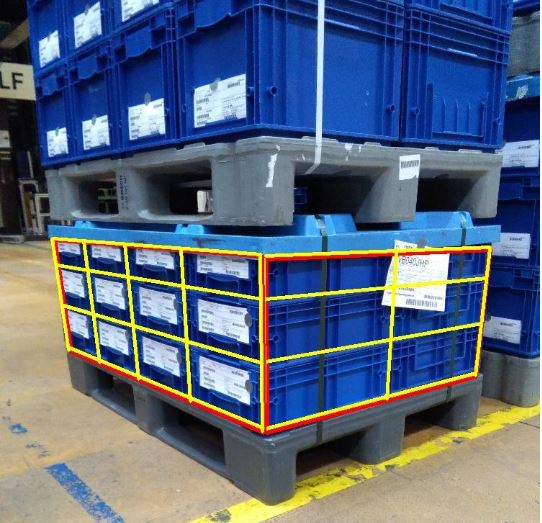}
	\caption{Illustration of the use-case of packaging structure recognition (taken from \cite{doerr2020psr})}
	\label{fig:use_case_illustration}
\end{figure}

The use-case of logistics packaging structure recognition aims at inferring the number, type and arrangement of standardized load carriers in uniform logistics transport units from a single image of that unit.
It is illustrated in Fig. \ref{fig:use_case_illustration}.
In an approach to design a robust solution to this task, we identified the recognition of two visible transport unit side faces, by finding the exact positions of their four corner points, as a reasonable sub-task \cite{doerr2020psr}.
Notably, our objects of interest, i.e. transport unit side faces, are of rectangular shape in real world.
As the perspective projection is the main component of the imaging transformation, we can assume, that transport unit side faces can be accurately segmented by four image pixel coordinates in regular images of logistics transport units.
Such assumptions are also valid in other logistics use-cases such as package detection or transport label detection.
The same holds for non-logistics applications like license plate or document recognition and other casess where objects of regular geometric shapes need to be segmented accurately to perform further downstream processing, like image rectifications or perspective transforms.

\begin{figure}[htb]
	\centering
	\includegraphics[trim=0 160 0 100,clip,width=0.35\textwidth]{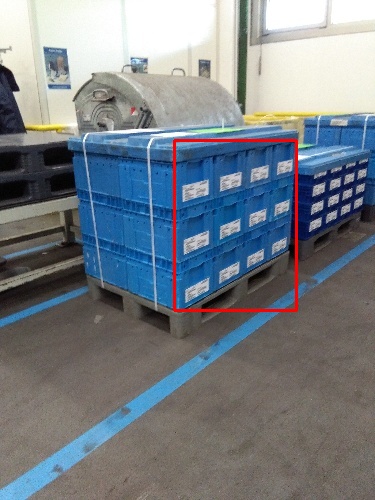}
	\includegraphics[trim=0 160 0 100,clip,width=0.35\textwidth]{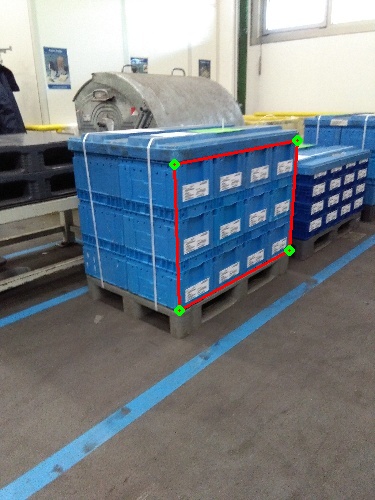}
	\caption{Sample annotations. Left: Bounding box. Right: Four vertices.}
	\label{fig:sample_annotations}
\end{figure}

To solve the challenge of detecting an object by finding a previously known number of feature points (e.g. four vertices), various approaches are thinkable.
For instance, the application of standard instance segmentation methods and adequately designed post-processing algorithms, simplifying the obtained pixel masks, may be a viable solution.
We aim to incorporate the geometrical a-priori knowledge into a deep-learning model by designing a convolutional neural network (CNN) detecting objects by four arbitrary feature points.
To achieve that, we build upon existing work by Law et al. \cite{law2018cornernet}, \cite{law2019cornernetlite}, enhancing the ideas of CornerNet.
CornerNet finds the two bounding box corners to solve the task of classic object detection.
We extend this idea to design a model detecting four arbitrary vertices of tetragonal shaped objects.
Fig. \ref{fig:sample_annotations} illustrates the difference between common bounding box object representations and our four-point based representations. 
The example image is taken from our use-case specific dataset. 
The objects indicated are transport unit faces which need to be precisely localized.

In this paper, we present a redesigned version of CornerNet, namely TetraPackNet, which segments objects by four object vertices instead of bounding boxes or pixel-masks.
We evaluate the approach on data concerning the use-case of logistics packaging structure recognition.
Notably, TetraPackNet cannot be baselined against its role model CornerNet, as it solves a fundamentally different task on differently annotated data.
Baselined against an existing solution to the sub-task of transport unit side corner detection, we show that TetraPackNet achieves improved results.
We observe that TetraPackNet is able to predict and localize tetragonal objects accurately.

The rest of the paper is organized as follows:
We summarize related work in Section \ref{sec:related_work}.
The model itself is explained in Section \ref{sec:method}.
Section \ref{sec:application} concerns the example application of logistics packaging structure recognition and the corresponding dataset.
We evaluate our approach in Section \ref{sec:experiments}.
Finally, Section \ref{sec:summary} concludes our work with a summary and outlook.

%% file: 2-relatedwork.tex
\section{Related Work}
\label{sec:related_work}

The primary use-case pushing our work is the one of logistics packaging structure detection, which we introduced in \cite{doerr2020psr}.
Apart from our own work, we are not aware of any other publications considering the same use-case.

As the number and frequency of publications regarding image object detection is enormous, we refer to dedicated survey papers as introductory material.
For instance, Wu et al. \cite{wu2020survey} or Liu et al. \cite{liu2020survey} give comprehensive overviews.

Our work builds on CornerNet \cite{law2018cornernet}, a recent work by Law et al., which aims to perform object detection without incorporating anchor boxes or other object position priors.
Instead, corner positions of relevant objects' bounding boxes are predicted using convolutional feature maps as corner heat maps.
Corners of identical objects are grouped based on predicted object embeddings.
This approach, which outperformed all previous one-stage object detection methods on COCO \cite{lin2014coco}, was further developed and improved by Duan et al. \cite{duan2019centernet} and Zhou et al. \cite{zhou2019objects}.
The follow-up work by Law et al. \cite{law2019cornernetlite}, CornerNet-Lite, introduced faster and even more accurate variations of the original CornerNet method.
These advancements of the original CornerNet are not in our scope, we build upon the original work \cite{law2018cornernet}.

Another approach relevant for this work, is the deep learning based cuboid detection by Dwibedi et al. \cite{dwibedi2016deep}.
In this work in the context of 3D-reconstruction, cuboid shaped, class agnostic objects are detected and their vertices are precisely localized.
We refrain from comparing to this work for several reasons, one of which is the requirement for richer image annotations (cuboid based, eight vertices per object).
Further, we do not aim to reconstruct 3D-models from our images but aim to classify and interpret intra-cuboid information.

%% file: 3-method.tex
\section{Method}
\label{sec:method}

We present a novel method for four-corner based object detection based on CornerNet \cite{law2018cornernet}, a recent work of Law et al.
Whereas in traditional object detection, object locations are referenced by bounding boxes (i.e. top left and bottom right corner position), we work with more detailed locations described by four independent corner points.
The resulting shapes are not limited to rectangles but comprise arbitrary tetragons, i.e. four-cornered polygons.

We use model, ground-truth and loss function designs very similar to those proposed in CornerNet \cite{law2018cornernet}.
These components, and especially our modifications for TetraPackNet, targeting tetragon-based object detection, are explained in the following sections.

\subsection{Backbone Network}
As suggested and applied by Law et al. \cite{law2019cornernetlite}, we use an hourglass network \cite{newell2016stackedhourglass}, namely Hourglass-54, consisting of 3 hourglass modules and 54 layers, as backbone network.
Hourglass networks are fully convolutional neural networks.
They are shaped like hourglasses in that regard, that input images are downsampled throughout the first set of convolutional and max pooling layers.
Subsequently, they are upsampled to the original resolution in a similar manner.
Skip layers are used to help conserve detailed image features, which may be lost by the network's convolutional downsampling.
In TetraPackNet's network design, two instances of the hourglass network are stacked atop each other to improve result quality.

\subsection{Corner Detection and Corner Modules}
Following the backbone network's hourglass modules, so-called corner modules are applied to predict precise object corner positions.
CornerNet utilizes two such corner modules to detect top-left and bottom-right corners of objects' bounding boxes.
Our architecture includes four corner detection modules for the four object corner types top-left, top-right, bottom-left and bottom-right.
It is important to note that we do not detect corners of bounding boxes, but vertices of tetragon-shaped objects.

Analogously to the original CornerNet approach, each corner module is fully convolutional and consists of specific corner pooling layers as well as a set of output feature maps of identical dimensions.
These outputs are corner heat maps, offset maps and embedding.
They each work in parallel on identical input information: the corner-pooled convolutional feature maps.

We shortly revisit CornerNet's specific pooling strategy.
It is based on the idea that important object features can be found when starting at a bounding box top left corner and moving in horizontal right or vertical bottom direction.
More precisely, by this search strategy and directions, object boundaries will be found by bounding box definition.
In CornerNet, max pooling is performed in the corresponding two directions for both bounding box corner types.
The pooling ouputs are added to one another and the results are used as input for corner prediction components.
The authors show the benefits of this approach in several detailed evaluations.
In our case, where precise object corners are predicted, instead of bounding box corners, one may argue that pooling strategies should be reconsidered.
Still, for our first experiments, we retain this pooling approach.

%The corner module's three output sets are explained in the following.
%For each distinguished corner type, i.e. top left, top right, bottom left and bottom right corners in our case, the model includes one heat map predicting positions where the probability for a corner of the respective type is high.
%As the resolution of the corner module's feature maps is lower than that of the original input image, additional location offsets are predicted for each potential corner candidate.
%To enable the assembly of four corresponding object corners to an object, embeddings are predicted for each corner.
%These embeddings aim to take such values that corners of the same object are as similar as possible, while those of corners of distinct objects differ significantly.
%The before-mentioned components corner pooling, corner heat map, offsets and embedding maps are combined to form a single corner prediction module.

%More precisely, each corner module is a set of fully convolutional layers including corner heat maps for each object category, two offset maps for horizontal and vertical offset and, in case of one-dimensional embeddings, one embedding map. 
%Moreover, we extended the CornerNet architecture to include four corner prediction modules instead of two.
%Additionally, TetraPackNet's corner prediction modules do not aim to detect bounding box corners, but vertices of tetragonal-shaped objects.

\subsection{Ground-truth}
\label{method:ground-truth}
Required image annotations are object positions described by the object's four corner points, i.e. top left, top right, bottom left, bottom right corner.
It is required that both right corners are further right as their counterparts and, equivalently, both top corners are further up as the corresponding bottom corners.
For each ground-truth object one single positive location is added to each of the four ground-truth heatmaps. 
To allow for minor deviations of corner detections from these real corner locations, the ground-truth heatmaps' values are set to positive values in a small region around every corner location.
As proposed by CornerNet, we use a Gaussian function centered at the true corner position to determine ground-truth heatmap values in the vicinity of that corner.

% To Do: Check comp. of radius for Gaussian etc.

\begin{figure}[htb]
	\centering
	\includegraphics[trim=15 0 15 0,clip,width=0.21\textwidth]{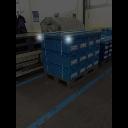}
	\includegraphics[trim=15 0 15 0,clip,width=0.21\textwidth]{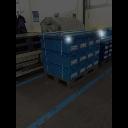}
	\includegraphics[trim=15 0 15 0,clip,width=0.21\textwidth]{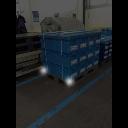}
	\includegraphics[trim=15 0 15 0,clip,width=0.21\textwidth]{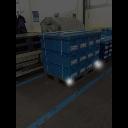}\vspace{0.02cm}\\
	\includegraphics[trim=15 0 15 0,clip,width=0.21\textwidth]{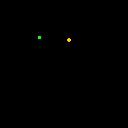}
	\includegraphics[trim=15 0 15 0,clip,width=0.21\textwidth]{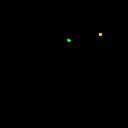}
	\includegraphics[trim=15 0 15 0,clip,width=0.21\textwidth]{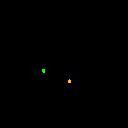}
	\includegraphics[trim=15 0 15 0,clip,width=0.21\textwidth]{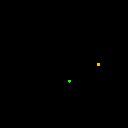}\\
	\footnotesize{\hspace{0.2cm} Top-Left \hspace{1.0cm} Top-Right \hspace{0.9cm} Bottom-Left \hspace{0.5cm} Bottom-Right}
	\caption{Example heatmaps. Top row: Groundtruth. Bottom row: Detected heats and color-encoded embeddings.}
	\label{fig:heatmaps}
\end{figure}

In Fig. \ref{fig:heatmaps} ground-truth and detected heatmaps, and embeddings are illustrated.
The top row shows, cross-faded on the original input image, the ground-truth heatmaps for the four different corner types.
There are two Gaussian circles in each corner type heatmap as there are two annotated ground-truth objects, i.e. two transport unit sides, in the image.
The bottom row shows TetraPackNet's detected heatmaps (for object type transport unit side) and embeddings in a single visualization:
Black regions indicate positions where the predicted heat is smaller than 0.1.
Wherever the detected heat value exceeds this threshold, the color indicates the predicted embedding value.
To map embedding values to colors, the range of all embeddings for this instance was normalized to the interval from 0 to 1.
Afterward Open CV's Rainbow colormap was applied \cite{bradski2008learning}.

\subsection{Loss Function}
The loss function used in training of our TetraPackNet model is structurally identical to that of CornerNet and consists of several components:
\begin{equation}
	L = L_{\text{det}} + w_{\text{off}} \cdot L_{\text{off}} + (w_{\text{pull}} \cdot L_{\text{pull}} + w_{\text{push}} \cdot L_{\text{push}})
\end{equation}

The only loss components we slightly modified to transition from CornerNet to TetraPackNet are the pull and push losses $L_{\text{pull}}$ and $w_{\text{push}}$.
These components are used to optimize the embedding values predicted at each potential corner location.
Modifications were made to account for the increased number of feature points to be detected.
Our version of pull and push loss are computed as follows
\begin{equation}
	\label{eq:pull_loss}
	L_{\text{pull}} = \frac{1}{N} \sum_{k=1}^{N} \sum_{i \in \{tl, tr, bl, br\} } {(e_{i}(k_{i}) - e(k))}^2
\end{equation}

\begin{equation}
	\label{eq:push_loss}
	L_{\text{push}} = \frac{1}{N(N-1)} \sum_{k=1}^{N} \underset{j\neq k}{\sum_{j=1}^{N}} \max{\{0, 1 - |e(k) - e(j)|\}}
\end{equation}

In our experiments, the loss component weights were set to $w_{\text{pull}} = w_{\text{push}} = 0.1$ and $w_{\text{off}} = 1.0$, as proposed by Law et al.

\subsection{Assembling Corner Detections to Objects}

Once corner positions and their embeddings are predicted, these predictions need to be aggregated to form tetragon object detections.
Compared to the CornerNet setup, this task is slightly more complex as each object is composed of four vertices instead of only two.
However, the original grouping implementation is based on Associative Embeddings \cite{newell2016associative}, which is suitable for multiple data points in general, i.e. more than two.
The same approach can be applied in our case.

To obtain an overall ranking for all detected and grouped objects, the four corner detection scores as well as the similarity of their embeddings, i.e. the corresponding pull loss values, are considered.
This final score for a detection $p$ of class $c$ consisting of four corners ${p_{tl}, p_{tr}, p_{bl}, p_{br}}$ is computed as

\begin{equation}
	\frac{1}{4} \sum_{i\in \{tl, tr, bl, br\}} h_i(c,p_i) + (e_i(p_i) - e(p))^2
\end{equation}

with $e(p)$ being the average embedding for a set of four corners as before.
Further, $h_i(c,p_i)$ denotes the predicted heat value for class $c$ and corner type $i \in \{tl, tr, bl, br\}$ at position $p_i$.

Additionally, we only allow corners to be grouped which comply with the condition, that right corners are further right in the image than their left counterparts.
Analogously bottom corners are required to be further down in the image than corresponding top corners.

%\subsection{Implementation}
%Our implementation of TetraPackNet is based on the open source CornerNet Lite implementation by the Princeton Vision \& Learning Lab \cite{princeton2020cornernetimplementation}.
%This Python implementation is based on PyTorch 1.0 and uses standard COCO evaluation packages (pycocotools).
%To apply the above-mentioned modifications to the original method, several components were added or re-implemented.
%These changes include ground-truth data handling and data sampling, CNN redesign, loss function, output decode function, and evaluation preparations.

%% file: 4-use-case-data.tex
\section{Use-Case and Data}
\label{sec:application}

\subsection{Logistics Unit Detection and Packaging Structure Analysis}
\label{sec:use_case}

Our work was developed in context of the logistics task of automated packaging structure recognition.
The aim of logistics packaging structure recognition is to infer the number and arrangement of a well standardized logistics transport unit by analyzing a single RGB image of that unit.
Fig. \ref{fig:use_case_illustration} illustrates this use-case.
To infer a transport unit's packaging structure from an image, the target unit, the unit's two visible faces and the faces of all contained load carriers are detected using learning-based detection models.
Several restrictions and assumptions regarding materials, packaging order and imaging are incorporated to assure feasibility of the task:
First of all, all materials like load carriers and base pallets are known. 
Further, transport units must be uniformly and regularly packed.
Each image shows relevant transport units in their full extend and in an upright orientation and in such a way, that two faces of each transport unit are visible.
The use-case and its setting, as well as limitations and assumptions are thoroughly explained in \cite{doerr2020psr} and \cite{doerr2020psr_fbv}.

We designed a multi-step image processing pipeline to solve the task of logistics packaging structure recognition.
The process' individual steps can be summarized as follows:

\begin{enumerate}
	\item Transport unit detection
	\item Transport unit side and package unit face segmentation
	\item Transport unit side analysis
	\item Information consolidation
\end{enumerate}

\begin{figure}[htb]
	\centering
	\includegraphics[width=0.18\textwidth]{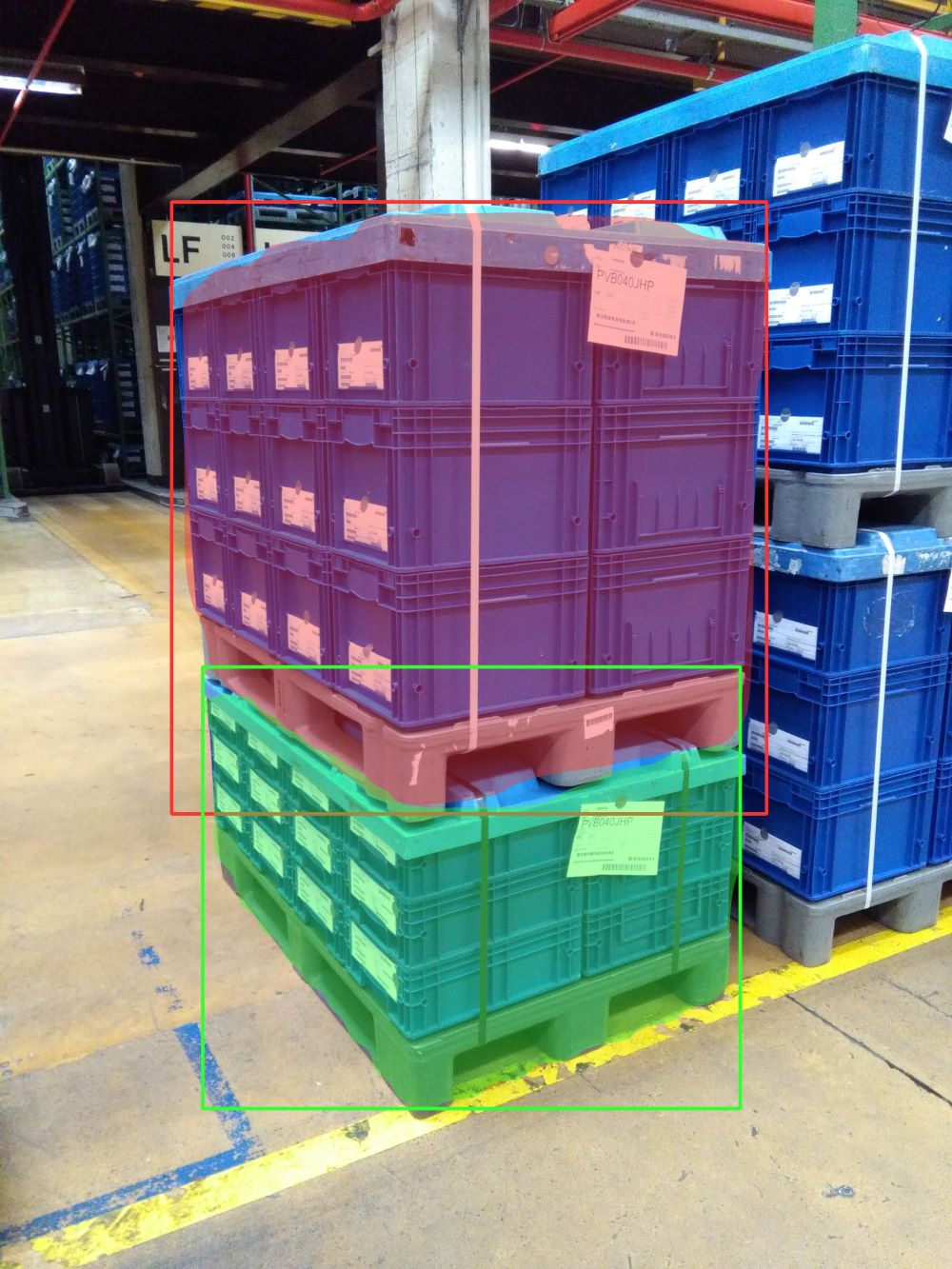}
	\includegraphics[width=0.235\textwidth]{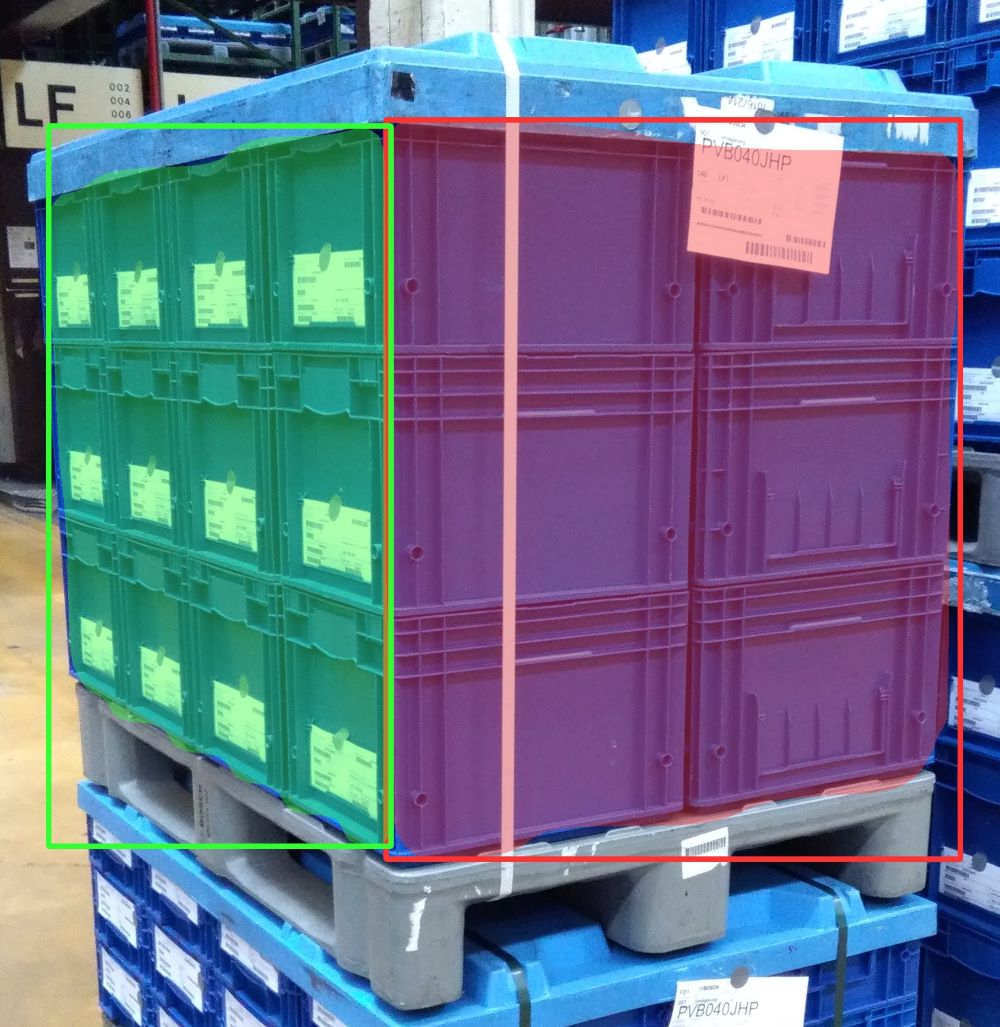}
	\includegraphics[width=0.198\textwidth]{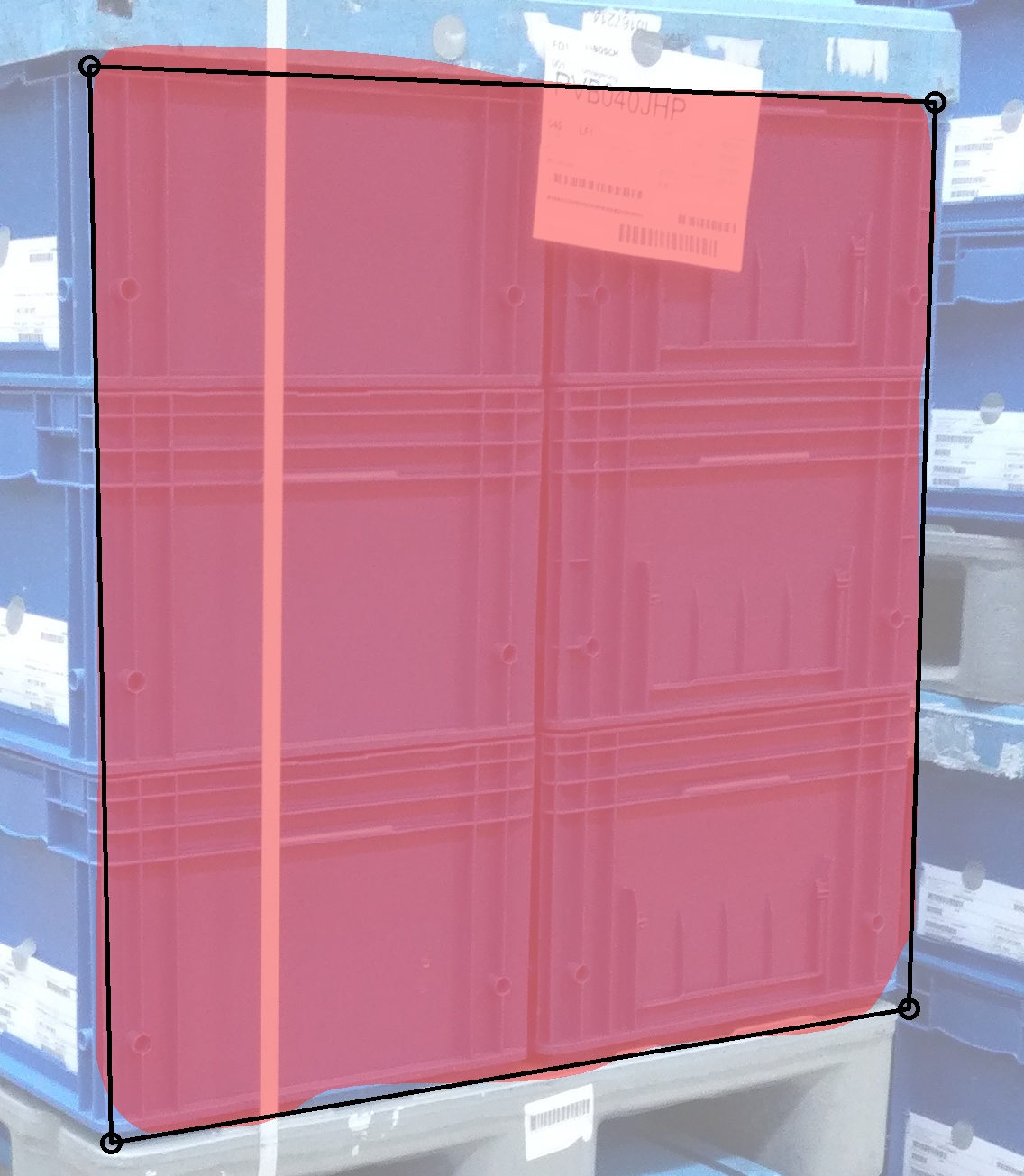} 
	\includegraphics[width=0.23\textwidth]{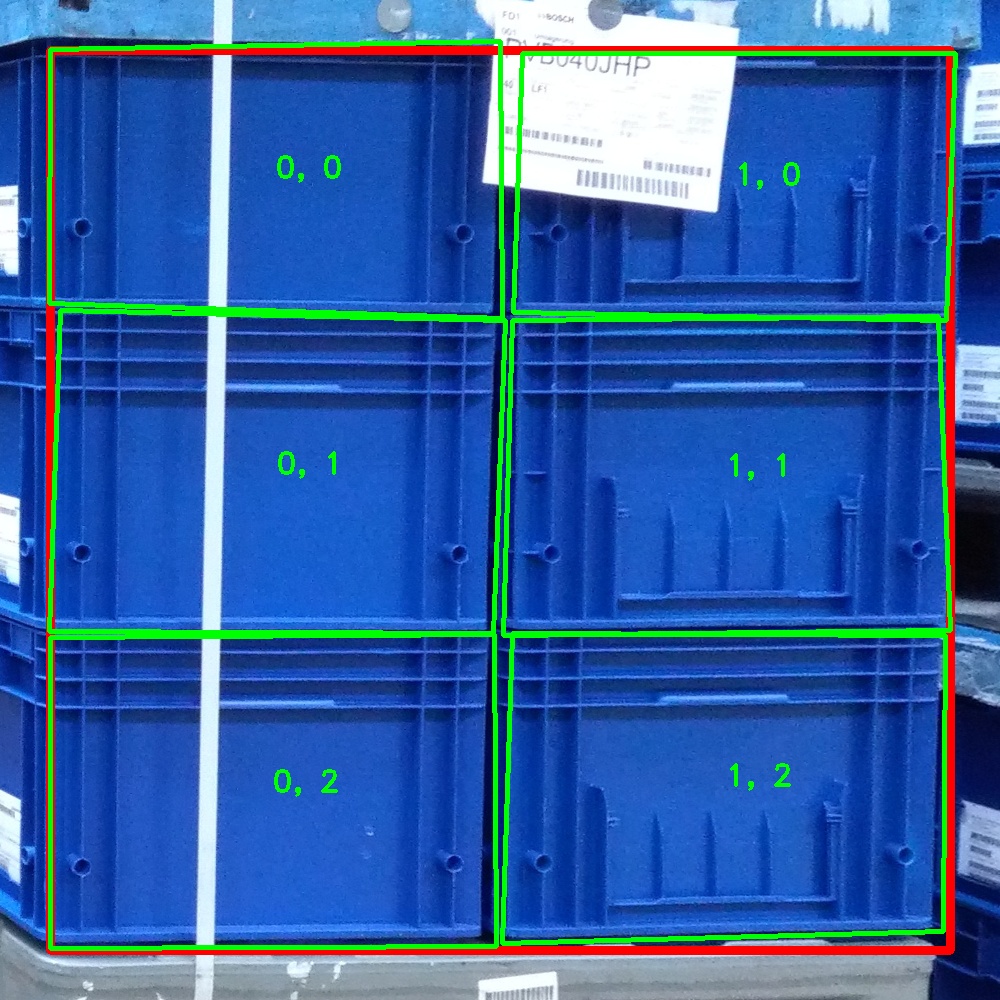} \\
	(a) \hspace{0.16\textwidth} (b) \hspace{0.18\textwidth} (c) \hspace{0.16\textwidth} (d) \\ \vspace{0.1cm}
	\caption{Method Visualization. (a) Transport unit identification. 
		(b) Transport unit side face segmentation. 
		(c) Side detection post-processing: Approximation of segmentation mask by four corner points. 
		(d) Rectified transport unit side.}
	\label{fig:method_overview}
\end{figure}

In step 1), whole transport units are localized within the image and input images are cropped correspondingly. (see Fig. \ref{fig:method_overview} (a)).
As a result, the input for step 2) is an image crop showing exactly one transport unit to be analyzed.
Subsequently, transport unit sides (and package units) are detected precisely.
This is illustrated in Fig. \ref{fig:method_overview} (b).
Step 3) aims to analyze both transport unit sides within the image.
This involves a rectification of the image's transport unit side region, in such a way to reconstruct a frontal, image boundary aligned view of each transport unit side.
To perform such a rectification, the precise locations of the side's four corner points are required.
Fig. \ref{fig:method_overview} (c) shows the rectified image patch of one transport unit side and illustrates package pattern analysis.
Each transport unit side is analyzed independently and its packaging pattern is determined.
In a last step, the information of both transport unit sides are consolidated.
The pipeline's overall results are the precise number and arrangement of packages for each transport unit within an image.

In the context of this work, we focus on step 2) of our overall packaging structure recognition pipeline.
Importantly, the simple detection and localization of these components represented by bounding boxes is not sufficient.
As we aim to rectify transport unit face image patches in a succeeding step, the precise locations of the four corners of both transport unit faces are required.
In our previous solution to this task, which was introduced in \cite{doerr2020psr}, a standard Mask R-CNN model \cite{he2017mask} is used to segment the transport unit faces in the image.
Afterwards, in a post-processing step, the segmentation masks, described by arbitrary polygons, are simplified to consist of four points only.
These points are refined by solving an optimization problem in such a way that the tetragon described by these four points has the highest possible overlap with the originally detected region.
Fig. \ref{fig:method_overview} (c) illustrates this four-corner approximation of segmentation masks.

In this work, we aim to replace the previously described procedure by TetraPackNet.
Instead of taking the detour via overly complex mask representations, we employ our specialized method TetraPackNet, which directly outputs the required four-corner based object representations.

\subsection{Specific Dataset}
\label{sec:data}

For training and evaluation of our TetraPackNet model, a custom use-case specific dataset of 1267 images was used.
The dataset was acquired in a German industrial plant of the automotive sector.
Each image shows one or multiple stacked transport units in a logistics setting.
The rich annotations for each image also include four-corner based transport unit side annotations.
The dataset was split in three sub-sets: training, validation and test data.
The test data set contains a handpicked selection of 163 images.
For a set of images showing identical transport units it was ensured that either all or none of these images was added to the test data set.
The remaining 1104 images were split into train and validation data randomly, using a train-validation ratio of 75-25.

%% file: 5-experiments.tex
\section{Experiments}
\label{sec:experiments}

In this section, we examine the performance of TetraPackNet.
We evaluate TetraPackNet on our use-case data and compare the results to baseline models and procedures, using standard and use-case specific metrics.
In a standard metric evaluation, the four-point regions found by TetraPackNet are compared to segmentation mask regions found by a baseline instance segmentation model.
This is done using the standard metric for the task of instance segmentation: mean Average Precision (mAP).
To evaluate TetraPackNet more specifically for our use-case, we compare its performance for the task of transport unit side detection with our previously implemented solution.
The previous solution, which was introduced in \cite{doerr2020psr} and described in section \ref{sec:use_case}, consists, again, of an instance segmentation model and adequate post-processing steps. 
In this evaluation, four-point represented regions are compared using use-case specific metrics.

\subsubsection{Setup}

Both models, TetraPackNet and the instance segmentation model, were trained for the single-class problem of transport unit side detection, as described in Section \ref{sec:use_case}.
In both cases, the same dedicated training, validation and tests splits were used.
Training and evaluation were performed on an Ubuntu 18.04 machine on a single GTX 1080 Ti GPU unit.
Two different training scenarios are evaluated for both models:
First, the models are trained to localize transport unit sides within the full images.
In a second scenario, the cropped images are used as input instead:
as implemented in our packaging structure recognition pipeline (see Section \ref{sec:use_case}), all images are cropped in such ways that each crop shows exactly one whole transport unit.
For each original image, one or multiple such crops can be generated, depending on the number of transport units visible within the image.
This second scenario is comparatively easier as exactly two transport unit sides are present in each image and the variance of the scales of transport units within the image is minimal.

\subsubsection{Training Details}

In both trainings, we tried to find training configurations and hyperparameter asssignments experimentally.
However, due to the high complexity of CNN training and its time consumption, an exhaustive search for ideal configurations could not be performed.
%Most likely, improvements are possible in both cases.
%Still, we consider the results presented in the following an affirmation of our proposed architecture TetraPackNet.
To achieve fair preconditions for both training tasks, the following prerequisites were fixed.
Both models were trained for the same amount of epochs:
The training of the Mask R-CNN baseline model included 200.000 training steps using a batch size of 1, whereas the TetraPackNet training included 100.000 training steps with a batch size of 2.
As backbone network, the Mask R-CNN model uses a standard Inception-ResNet-v2 \cite{szegedy2017inception} architecture.
Input resolution for both models was limited to 512 pixels per dimension. 
Images are resized such that the larger dimension measures 512 pixels and aspect ratio is preserved. 
Subsequently, padding to quadratic shape is performed.
In both trainings, we considered similar image augmentation methods: random flip, crop, color distortions and conversion to gray values.
Of these options, only the ones yielding improved training results were retained.

\subsection{Standard Metric Results}

First of all, we compare TetraPackNet to a model performing classic image instance segmentation, a more complex task than is solved by our novel model TetraPackNet.
Note that such a comparison only makes sense on data regarding four-point based object detection as TetraPackNet does not aim to solve the general task of instance segmentation.
Still this comparison is meaningful as instance segmentation models are capable of detecting arbitrary shapes, including tetragonal ones.
Additionally, we are not aware of other models performing four-point object detection which could be used as baseline methods.

As standard evaluation metric, the COCO dataset's \cite{lin2014coco} standards are used. 
We report average precision (AP) at intersection over union (IoU) threshold of 0.5 ($AP_{0.5}$), 0.75 ($AP_{0.75}$) and averaged for ten equidistant IoU thresholds from 0.5 to 0.95 ($AP$).
Evaluations are performed on our dedicated 163-image test dataset.
Table \ref{tab:exp1_results} shows the corresponding results.

\begin{table}
	\centering
	\caption{Evaluation results for the whole image scenario.}
	\begin{tabular}[c]{|l||p{0.1\textwidth}||p{0.1\textwidth}|p{0.1\textwidth}|p{0.1\textwidth}|}
		\hline
		\textbf{Model} & \textbf{$AP$} & \textbf{$AP_{0.5}$} & \textbf{$AP_{0.75}$} & \textbf{$AP_{0.9}$} \\
		\hline
		Mask R-CNN & 58.7 & 87.0 & 66.5 & 16.5 \\
		\hline
		TetraPackNet & 75.5 & 83.6 & 83.5 & 66.7 \\
		\hline
	\end{tabular}
	\label{tab:exp1_results}
	
	\vspace{0.3cm}
	\caption{Evaluation results for the cropped image scenario.}
	\begin{tabular}[c]{|l||p{0.1\textwidth}||p{0.1\textwidth}|p{0.1\textwidth}|p{0.1\textwidth}|}
		\hline
		\textbf{Model} & \textbf{$AP$ }& \textbf{$AP_{0.5}$} & \textbf{$AP_{0.75}$} & \textbf{$AP_{0.9}$} \\
		\hline
		Mask R-CNN & 80.3 & 98.9 & 91.5 & 54.6 \\
		\hline
		TetraPackNet & 91.1 & 96.0 & 95.0 & 85.2 \\
		\hline
	\end{tabular}
	\label{tab:exp2_cropping_results}
\end{table}

\begin{figure}[htb]
	\centering
	\includegraphics[width=1\textwidth]{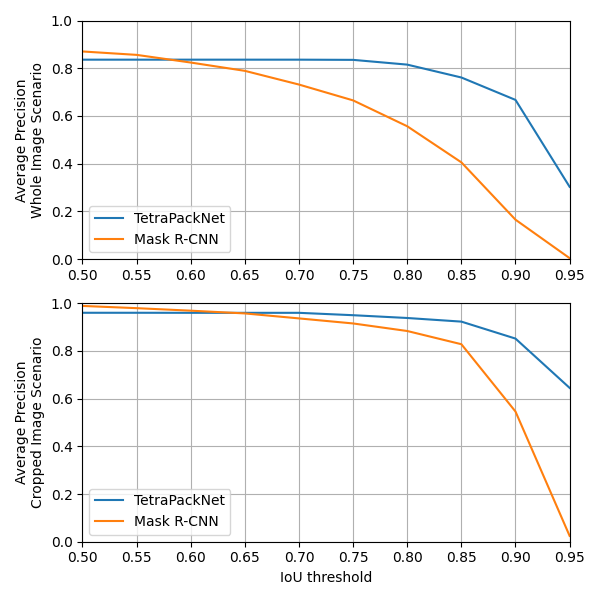}
	\caption{Average precisions at different IoU thresholds for TetraPackNet and Mask R-CNN baseline model. Left: Whole image scenario. Right: Cropped image scenario.}
	\label{fig:ap_plot}
\end{figure}

Considering only the values at the lowest IoU threshold examined (0.5), the baseline Mask R-CNN outperforms TetraPackNet by visible margins:
Mask R-CNN's $AP_{0.5}$-value is 3.4 points higher than that of TetraPackNet (87.0 vs. 83.6).
However, as the IoU threshold for detections to be considered correct increases, TetraPackNet gains the advantage.
When regarding performance values at IoU threshold 0.75 instead, TetraPackNet achieves a significantly higher precision of 83.5, compared to 66.5 for Mask R-CNN.
This observation can be expanded for the average precision scores at higher IoU thresholds: 
TetraPackNet begins to gain advantage over our baseline method as detection accuracy requirements increase.
This is illustrated and visible in the top-part of fig. \ref{fig:ap_plot}, which visualizes the same evaluation results shown in table \ref{tab:exp1_results}.

Very similar observations can be made for the cropped image scenario:
TetraPackNet clearly outperforms the reference model Mask R-CNN when high accuracy is required.
The corresponding evaluation results are shown in table \ref{tab:exp2_cropping_results}, and in the bottom part of Fig. \ref{fig:ap_plot}.

Overall, the results suggest that TetraPackNet does not detect quite as many ground-truth transport unit sides as our Mask R-CNN baseline model on a low accuracy basis.
At the same time, the predictions made by TetraPackNet appear to be very precise as average precision steadily remains on a high level as IoU accuracy requirements are increased.
For our use-case of packaging structure recognition, these are desirable conditions, as our processing pipeline requires very accurate four-point based transport unit side predictions.

\subsection{Use-Case Specific Results}

Within our use-case of packaging structure detection, we aim to localize each transport unit side by four corner points.
In this section, we baseline our results against our previous approach to transport unit side detection, which consists of a segmentation model and a succeeding post-processing procedure. 
The former is a Mask R-CNN model with an Inception-ResNet-v2 backbone network, as described above. 
The latter is necessary to obtain the required four-point object representations from the segmentation model's output masks.
This is performed by solving a suitable optimization model, which outputs four vertices giving the best approximation of a segmentation masks in terms of region overlap.
Input to the task in our image processing pipeline, and for these evaluation, are cropped images showing exactly one full transport unit.

To investigate the performance of TetraPackNet for our specific use-case, other metrics than standard COCO Average Precision are needed.
As evaluation criteria two different metrics are computed, both based on the standard value of intersection over union (IoU).
For each ground-truth transport unit side, the IoUs with its assigned detection, represented by four vertices, is computed.
Thereby, only the two highest-ranked detections of each method are considered, as there are exactly two transport unit sides in each image.
Detections are assigned from left to right, based on the a-priori knowledge, that one transport unit side is clearly positioned further left than the other one.
Only for cases where less than one transport unit side was detected by the methods under consideration, an IoU-based assignment, with IoU threshold 0.5, is performed.
If no detection was matched to a ground-truth object, the IoU value for this side is set to 0.

The first metric we compute is an overall accuracy value. 
We assume a ground-truth transport unit side to be detected correctly if it has an IoU of at least 0.8 with its assigned detection.
The accuracy of transport unit side detection is equal to the percentage of annotated sides detected correctly.

The second evaluation value we compute, is the average IoU for correctly assigned detections. 
For the use-case at hand this is a reasonable and important assessment, as the succeeding packaging structure analysis steps rely on very accurate four-corner based transport unit side segmentations \cite{doerr2020psr}.
Table \ref{tab:exp3_usecase_results_cropping} shows the corresponding results.

Overall, this use-case specific evaluation requiring a high detection accuracy is dominated by our novel approach TetraPackNet.
The latter achieves higher rates of high-accuracy transport unit side detections:
TetraPackNet correctly detects 95.7\% of transport unit sides, whereas the baseline method only achieves 86.6\% in this metric.
Additionally, if only considering sufficiently accurate detections of both models, the average IoU of the detections output by TetraPackNet was significantly higher (0.05 IoU points on average) than for those of our baseline method.
We deduce the suitability of TetraPackNet for our application.

\begin{table}
	\centering
	\caption{Use-case specific evaluation results on our 163-image test dataset.}
	\begin{tabular}[c]{|l|r|r|}
		\hline
		\textbf{Model} & \textbf{Accuracy} &  \textbf{Average IoU} \\
		&& (Positives only) \\
		\hline
		Mask R-CNN \& post-processing & 86.6 & 0.908 \\
		\hline
		TetraPackNet & 95.7 & 0.958 \\
		\hline
	\end{tabular}
	\label{tab:exp3_usecase_results_cropping}
\end{table}

%% file: 6-conclusion.tex
\section{Summary and Outlook}
\label{sec:summary}

\subsection{Results Summary}
We presented TetraPackNet, a new detection model outputting objects based on a novel four-vertex representation.
We showed that the use of such a specialized model, instead of an instance segmentation model employing more complex object representations, can lead to superior results in cases where corresponding object representations are reasonable and necessary.
This is the case in our presented use-case of logistics packaging structure recognition, but may also apply to numerous other tasks, as for instance document or license plate detection.

We trained and evaluated TetraPackNet on our own use-case specific dataset.
For the dedicated task, the observed results indicate a higher accuracy (superior by 9 percentage points) compared to a previous approach involving a standard Mask R-CNN model and suitable post-processing.

\subsection{Future Work}
The applicability of TetraPackNet to our use-case of logistics packaging structure recognition will be evaluated further.
First of all, we plan to apply the model to package unit detection.
This task is additionally challenging as, in general, a large number of densely arranged package faces of very similar appearance need to be detected.
On the other hand, a-priori knowledge about the regular package arrangement might allow for specific corner detection interpretations and even interpolations (e.g. in case of single missing corner detections).
Additionally, we plan to extend TetraPackNet to specialized detection tasks including even more than four corner or feature points.
In our case of logistics transport unit detections, a lot of a-priori knowledge about object structure, shape and posture is given.
This can be exploited by integrating specific transport unit templates into the detection model:
one can define multiple additional characteristic points (e.g. base pallet or packaging lid corners) which are to be detected by a highly specialized deep learning model.

Further, we will investigate different algorithmic and architectural choices.
As mentioned before, corner pooling strategies adopted from CornerNet might not be ideal for TetraPackNet and its applications.
Thus, for instance, experiments with different corner pooling functions will be performed.

At the same time, TetraPackNet is not necessarily limited to the use-case of packaging structure recognition or logistics in general.
We plan to affirm our positive results by evaluating TetraPackNet on additional datasets of different use-cases.